\documentclass{article}

\usepackage[T2A, T1]{fontenc} 
\usepackage[utf8]{inputenc}
\usepackage[russian, english]{babel}

\usepackage{amsmath}
\usepackage{amssymb}
\usepackage{mathrsfs}
\usepackage{stmaryrd}

\usepackage{libertine}
\usepackage[libertine]{newtxmath}
\usepackage{microtype} 

\usepackage[margin=1in]{geometry}
\usepackage[round]{natbib}
\usepackage{graphicx}
\usepackage{array}
\usepackage{float}
\usepackage[table]{xcolor}
\usepackage{tabularx}
\usepackage{booktabs}
\usepackage{multirow}
\usepackage{textcomp}
\usepackage{fancyhdr}
\usepackage{setspace}
\usepackage{endnotes}
\usepackage{url}
\usepackage{authblk}
\definecolor{darkgreen}{rgb}{0,0.5,0}
\definecolor{darkred}{rgb}{0.6,0.0,0}
\definecolor{teal}{rgb}{0.00, 0.62, 0.45}
\definecolor{coral}{rgb}{0.83, 0.37, 0.00}
\newcommand{\usc}{\underline{\hspace{0.4cm}}\ }
\newcommand{\angles}[1]{\langle {#1} \rangle\ }
\newcolumntype{L}{>{\raggedright\arraybackslash}X}

\newcommand{\heatmap}[1]{%
    \ifdim#1pt<50pt\cellcolor{gray!20}#1\%\else%
    \ifdim#1pt>72pt\cellcolor{blue!50}#1\%\else%
    \ifdim#1pt>60pt\cellcolor{blue!20}#1\%\else%
    \ifdim#1pt>50pt\cellcolor{blue!10}#1\%\fi\fi\fi\fi}

\usepackage{gb4e}

\begin{document}

\author[1]{Imry Ziv}
\author[2]{Nur Lan}
\author[2]{Emmanuel Chemla}
\author[1]{Roni Katzir}

\affil[1]{Tel Aviv University
}
\affil[2]{École Normale Supérieure}

\title{Large Language Models as Proxies for Theories of Human Linguistic Cognition\thanks{
We thank Danny Fox, Fred Landman, Roger Levy, Eyal Marco, and audiences at Leibniz ZAS, MIT, and TAU, as well as the reviewers and editors at JLM. RK has been supported by ISF grant \#1083/23 and by the Alexander von Humboldt Foundation. This project was provided with computer and storage resources by GENCI at IDRIS thanks to the grant AD011013783R2 on the supercomputer Jean Zay’s V100 and CSL partitions.}}

\maketitle              %

\begin{abstract}
We consider the possible role of current large language models (LLMs) in the study of human linguistic cognition. We focus on the use of such models as proxies for theories 
of cognition that are relatively linguistically-neutral in their representations and learning but differ from current LLMs in key ways. 
We illustrate this potential use of LLMs as proxies for theories of cognition in the context of two kinds of questions:  (a) whether the target theory accounts for the acquisition of a given pattern from a given corpus; and (b) whether the target theory makes a given typologically-attested pattern easier to acquire than another, typologically-unattested pattern. For each of the two questions we show, building on recent literature, how current LLMs can potentially be of help, but we note that at present this help is quite limited.
\end{abstract}


\maketitle
\section[Background: LLMs, HLC, inference to best explanation]{Background: LLMs, HLC, and the inference to the best explanation}
\label{sec:bg}

Being part of empirical science, the study of HLC proceeds through inference to the best explanation: explicit theories are evaluated in light of the observations, and those theories that provide the best explanations of the data are preferred to the other contenders. The theories coming out of generative linguistics over the past 70 years or so are routinely evaluated in this way. These theories are often strongly linguistically biased, in the sense that they make it relatively easy to represent and learn knowledge of the kind found across languages and hard or impossible to represent or learn many other patterns.
Let us use $H_1$ to refer to some concrete instantiation of the generative approach. We will return to the biases embodied by the architecture and representations of $H_1$ in the following sections.

More recently, \citet{Piantadosi:2023} has suggested that current LLMs 
should be treated as explicit theories of HLC that should be taken seriously as part of this evaluation. Following \citealt{FoxKatzir:2024} we refer to this new approach to HLC as the \textit{LLM Theory}, and we discuss it in Section \ref{sec:llm:th}, through a concrete instantiation we will call $H_2$. Perhaps in light of the glaring inadequacies of the LLM Theory, some recent literature, including \citealt{WilcoxFutrellLevy:2023} and \citealt{MahowaldIvanovaBlankKanwisherTenenbaumFedorenko:2024}, has stopped short of explicitly endorsing the LLM Theory while still taking the performance of current LLMs to challenge central claims within theoretical linguistics. It is conceivable that this more recent literature is still meant to support the LLM Theory, in which case the objections to that theory, reviewed below, still apply. But it is also possible that at least some of these works aim at a more nuanced view on the role of LLMs in studying HLC, one in which the performance of current models can be taken to challenge generative views and support a theory of HLC $H_3$ that is not as strongly linguistically-biased as $H_1$ but is also very different from current models themselves, and therefore different from $H_2$. Our goal in this paper is to articulate this more nuanced view and to make a preliminary attempt to evaluate it in light of current evidence. We also hope to encourage adherents of this view to make explicit the target $H_3$ they have in mind.

The idea, as we understand it, is the following. If one wishes to challenge claims for strong linguistic biases, as in $H_1$ mentioned above, and to defend a more linguistically-neutral alternative $H_3$, one would need to address the arguments that theoretical linguistics has provided for $H_1$ from acquisition and from the typology, among other things. But doing so directly can be difficult. For one thing, $H_3$ might make it hard to test acquisition predictions based on developmentally-realistic corpora. It can also be hard to evaluate the ability of $H_3$ to predict typological patterns. It is at this point that current LLMs can be helpful. While these models might be quite different from $H_3$, they too lack the strong linguistic biases that theoretical linguistics has argued for, so their successful performance on tasks related to learning and typology might lend support to the possibility that $H_3$ will succeed on these tasks despite not having the strong linguistic biases that theoretical linguistics has argued for. In this way, current LLMs can serve as proxies for $H_3$. We refer to this idea as the \textit{Proxy View}:

\begin{exe}
\ex{\label{def:prox}The Proxy View of LLMs in HLC: current LLMs are not themselves good theories of HLC, but their performance is indicative of the success of some other theory of HLC, presumably more linguistically-neutral than the generative approach, and can inform the inference to the best explanation.
}
\end{exe}

Since the literature that promotes the Proxy View generally does not make its supported theory explicit, our ability to evaluate it at present is limited. We will therefore focus on methodological considerations that pertain to the evaluation of the target theory once it is provided and show how LLMs might bear on this evaluation. We show that, at least at present and with respect to the phenomena considered here, LLMs lend no support to the Proxy View.

\subsection{The LLM Theory and its inadequacy}
\label{sec:llm:th}

As mentioned, \citet{Piantadosi:2023} argues for the LLM Theory of HLC. That is, he suggests that current LLMs, taken as real theories of HLC, are a better explanation of the known observations than anything coming out of the generative tradition. This claim is incorrect, for reasons discussed in detail in the literature and summarized briefly here. Following \citealt{Katzir:2023} and \citealt{FoxKatzir:2024} we structure our brief discussion of the failure of the LLM Theory around three key issues: (a)~competence vs.\ performance, (b)~correctness vs.\ probability, and (c)~learning and representations.%
\footnote{See  \citealp{KodnerPayneHeinz:2023}, \citealp{MoroGrecoCappa:2023}, \citealp{RawskiBaumont:2023}, and others for further critical discussion of the LLM Theory.}

\textbf{Competence vs.\ performance.} Humans struggle with center embedding even at very modest depths of embedding, as in ``The mouse that the cat that the dog chased bit died''. A naive account might attribute this to the grammar, part of the competence of speakers: perhaps human grammars rule out center embedding beyond a certain (low) threshold. A much better explanation, going back to \citealt{Yngve:1960} and \citealt{MillerChomsky:1963}, attributes the difficulty with center embedding not to the competence of speakers but rather to their performance, typically by attributing it to limitations of their working memory. Among other things, the account in terms of performance explains why center embedding is hard, as seen in reaction times and various errors, as well as from introspection, rather than being just unacceptable. Unacceptability due to the grammar alone, as in ``John ate apple the'', does not share this property; it is simply judged to be bad. The account in terms of the distinction between competence and performance also explains why the unacceptability of center embedding is gradient, growing with each level of embedding, and why it can vary with certain factors (including with the use of pronouns instead of referring nouns in the most embedded position). Again, the account in terms of competence alone has no explanation for these facts. 



Differently from common generative proposals, the design and behaviour of current LLMs does not suggest a meaningful distinction between competence and performance of the kind discovered in humans: there is no reason to think that more (or less) time and working memory would make them change their behaviour on center embedding or agreement. This leaves current LLMs with no good explanation for the increasing difficulty of sentences with increasing levels of center embedding or for the other observations that have been taken to support the distinction between competence and performance.

\textbf{Correctness vs.\ likelihood.}  Humans recognize some sentences as correct but unlikely (one example again is multiple center embedding as in the example above), and they recognize some other sentences as incorrect but likely (for example, so-called agreement attraction, as in the incorrect ``The keys to the cabinet is on the table'', where \textit{is} appears in the singular despite a plural subject \textit{keys}, presumably because of the linear proximity of the singular \textit{cabinet}). The significance of the distinction between correctness and likelihood for linguistics was pointed out by \citet{Chomsky:1957} and discussed further by \citet{Berwick:2018}, \citet{SprouseYankamaIndurkhyaFongBerwick:2018}, and others.

LLMs provide straightforward information regarding their representation of likelihood: given a prefix, they provide a probability distribution over the next token. But they do not, at present, provide information about correctness as distinct from likelihood, and it remains unclear whether they even have suitable representations to support this distinction.\footnote{A potentially important area of research concerns understanding the inner workings of LLMs, particularly through the study of mechanistic interpretation. See \citealt{Lakretz:2019}, \citealt{Ravfogel:2021}, \citealt{Rai:2024} and \citealt{LopezOtal:2025}, among others. Such investigations have the potential of making the inner workings of LLMs less opaque, and it is conceivable that they will eventually uncover a notion of correctness within LLMs.}

We note that the literature often compares the likelihood assessments of LLMs to acceptability judgments of speakers. Such, for example, is the methodology of works that evaluate the preferences of LLMs within minimal pairs, including \citealt{LinzenDupouxGoldberg:2016}, \citealt{BernardyLappin:2017},  \citealt{ChowdhuryZamparelli:2018},  \citealt{GulordavaBojanowskiGraveLinzenBaroni:2018}, \citealt{KuncoroDyerHaleYogatamaClarkBlunsom:2018}, \citealt{MarvinLinzen:2018}, \citealt{WilcoxLevyMoritaFutrell:2018,WilcoxLevyFutrell:2019,WilcoxFutrellLevy:2023}, \citealt{BhattacharyaSchijndel:2020}, \citealt{Chaves:2020},   \citealt{WarstadtParrishLiuMohananeyPengWangBowman:2020}, \citealt{HuebnerSulemCynthiaRoth:2021}, \citealt{OzakiYurovskyLevin:2022},  \citealt{yedetore:2023}, and \citealt{LanChemlaKatzir:2024a}. This methodology can be informative for particular purposes and under particular conditions, and in Section \ref{sec:aps} we adopt it ourselves, but this should not be taken to imply that the relevant likelihood assessments correspond to correctness. As mentioned, the two notions are distinct.\footnote{A different method that might appear to offer insight into a notion of correctness or grammaticality in LLMs is prompting, often through interfaces such as ChatGPT. This roughly corresponds to comparing the distribution over `yes' and `no' given a prefix such as ``Is the following grammatically correct?'' + \texttt{<target-sentence>}. (The precise evaluation depends on the particular prompt that is used and can vary further depending on the specifics of the relevant chat interface.) This is often a convenient way to anecdotally illustrate a general point, by making an LLM appear to engage in reporting its acceptability assessment of the target sentence. While some work, including \citealt{DentellaGuntherLeivada:2023} and \citealt{QiuDuanCai:2025}, uses prompting on a larger scale, aiming at making general points rather than just illustration, we find this use hard to interpret. In any event, we know of no particular reason to think that prompting provides access to any internal notion of correctness within the LLM (if such a notion even exists).}

\textbf{Learning and representations.} The inductive leaps that humans make, sometimes based on very little data, suggest specific representations and biases. This is the essence of so-called \textit{arguments from the poverty of the stimulus} (APSs). A particularly well-known APS, originally outlined by \citet{Chomsky:1971} and discussed in detail in much later work (see \citealt{BerwickPietroskiYankamaChomsky:2011} for a more recent assessment), concerns hierarchical structure, often stated in terms of constituency.\footnote{As a reviewer notes, this APS concerns several other syntactic notions in addition to hierarchical structure. For presentational purposes we focus the current discussion on hierarchical structure.} This is often illustrated with the formation of yes-no questions, as in (\ref{sai}). The generalization that English-learning children reach is that such questions are formed by fronting the auxiliary that is structurally highest, as in (\ref{sai:good}). They do not seem to seriously consider an alternative generalization that says that the fronted auxiliary is the one that is first linearly (which would yield the incorrect (\ref{sai:bad})), even though the data that children are typically exposed to has been argued to be unhelpful in making the choice.

\begin{exe}
\ex{\label{sai}[The boy who is singing] can dance
\begin{xlist}
\ex[]{\label{sai:good}Can [the boy who is singing] \usc\ dance?}
\ex[*]{\label{sai:bad}Is [the boy who \usc\ singing] can dance?}
\end{xlist}
}
\end{exe}

Generative theories of HLC typically derive this inductive leap by assuming that children are born with a combination of representations and learning biases that favour dependencies that rely on hierarchical structure over dependencies that rely on linear order (in some approaches by preventing the latter from being represented in the first place). Current LLMs, on the other hand, seem to provide no basis for this inductive leap. In fact, recent work by \citet{LanGeyerChemlaKatzir:2022,LanChemlaKatzir:2024b} casts doubt on the ability of networks trained by current methods to even acquire the basic hierarchical notions, let alone to explain the preference for dependencies that rely on these notions over dependencies that rely on linear order. Not surprisingly, then, \citet{yedetore:2023} find evidence suggesting that at least some neural networks fail to acquire the correct hierarchically-based generalization from their training data. We will return to APSs in Section \ref{sec:aps}, where we will consider the possible role of LLMs in reasoning about the inductive leaps predicted by non-LLM theories of HLC.

Evidence about the representations and learning biases of humans also comes from the highly skewed distribution of linguistic patterns cross-linguistically. We can again consider hierarchical structure, following \citealt{FoxKatzir:2024}, though we emphasize that a very wide range of cross-linguistic patterns have been discussed in the linguistic literature. We just mentioned that biases that refer to hierarchical structure are part of the best explanation of the competence of English speakers, noting evidence from the formation of yes-no questions. Notions of hierarchical structure are central to the explanation of many other aspects of English, including center-embedding and agreement mentioned earlier, wh-movement discussed in Section \ref{sec:aps} below, the computation of meaning, and more. But even if one ignores the APSs for hierarchical structure, it seems clear that hierarchical structure is not just an accident of actual English: it is part of the best explanation of almost all languages that have been studied closely, which suggests that humans are born with representations and learning biases that favour hierarchical structure. Again, the representations and learning biases of current LLMs might not even be able to support the acquisition of the relevant notions of hierarchical structure, let alone explain the cross-linguistic prevalence of these notions. We will return to cross-linguistic evidence in Section \ref{sec:typ}, where we will consider the possible role of LLMs in reasoning about the typological patterns predicted by non-LLM theories of HLC.

\subsection{LLMs as proxies for good learners}
But if the LLM Theory is discredited, does this mean that current LLMs are irrelevant to the scientific study of HLC? As mentioned above, the Proxy View explains why the answer is no: in principle current LLMs can serve as proxies for theories of a third kind, theories that involve relatively linguistically-neutral representations and learning (and are different in this regard from typical generative theories) but that are not vulnerable to the most immediate arguments against the LLM Theory. Table~\ref{tab:theories} schematically compares the three kinds of theories of HLC that we discuss here: a concrete generative theory, $H_1$; a concrete instantiation of the LLM Theory, $H_2$; and a concrete, relatively linguistically-neutral, $H_3$, possibly very different from current LLMs, for which these LLMs might serve as proxies. The Proxy View, as stated in (\ref{def:prox}) above, uses current LLMs as tools for evaluating $H_3$, in comparison with $H_1$ and as part of the inference to the best explanation in light of the existing observations and arguments.

At this point we should mention again an immediate obstacle to evaluating the Proxy View: while recent literature, including  \citealt{WilcoxFutrellLevy:2023} and \citealt{MahowaldIvanovaBlankKanwisherTenenbaumFedorenko:2024}, suggests that the performance of LLMs supports linguistically-neutral theories of HLC, this literature does not provide much by way of specifics. Until such specifics of $H_3$ are provided, and until $H_3$ is shown to be a realistic candidate for the best explanation in light of the existing literature, expressions of hope about $H_3$ remain just that. 

Still, we can outline some methodological considerations regarding the possible role of current LLMs as proxies for $H_3$ once the specifics of $H_3$ are made sufficiently explicit. Here we consider two such uses of LLMs as proxies for $H_3$. Following \citealt{LanChemlaKatzir:2024a} we use LLMs as tools for reasoning about whether $H_3$ predicts a given inductive leap based on a given amount of linguistic exposure. And following \citealt{KalliniPapadimitriouFutrellMahowaldPotts:2024} we use LLMs as tools for reasoning about whether $H_3$ derives a learning asymmetry that might be part of an explanation of a typological asymmetry.

\begin{table}
\centering
\renewcommand{\arraystretch}{1.5}
\small
\resizebox{\columnwidth}{!}{%
\begin{tabular}{l|ccc}
\hline
\textbf{Criterion} & \textbf{Generative~linguistics} & \textbf{LLM Theory} & \textbf{Proxy View} \\
\hline
Linguistic bias & \cellcolor{green!25} Strong & \cellcolor{red!25} Weak & \cellcolor{green!25} Intermediate \\
Competence vs. performance & \cellcolor{green!25} $\checkmark$ & \cellcolor{red!25} $\times$ & \cellcolor{green!25} $\checkmark$ \\
Grammaticality $\neq$ likelihood & \cellcolor{green!25} $\checkmark$ & \cellcolor{red!25} $\times$ & \cellcolor{green!25} $\checkmark$ \\
Explanation for learning asymmetries &
\cellcolor{green!25} $\checkmark$ &
\cellcolor{red!25} $\times$ &
\cellcolor{green!25} $\checkmark$ \\
Concrete instantiation &
\cellcolor{green!25} $H_1 $&
\cellcolor{red!25} 
$H_2$ &
\cellcolor{green!25} $H_3$ \\
\hline
\end{tabular}
}

\caption{Comparison of the three theories outlined above on key issues, following~\citealt{FoxKatzir:2024}.} 
\label{tab:theories}
\end{table}

\section{Large language models and alignment with the stimulus}
\label{sec:aps}

Consider then an $H_3$ that is neutral with respect to some linguistic property $PROP$ for which generative theories such as $H_1$ include a strong linguistic bias. A potential argument against $H_3$ and in favour of $H_1$ might take the form of an APS: if humans show knowledge of $PROP$ after exposure to a corpus that seems too impoverished to support the learning of $PROP$ by a linguistically-neutral learner, this will appear to support the strongly biased $H_1$. The problem is that it is generally very hard to determine what a given learner might acquire from a given corpus, even for a theory such as $H_{3}$ for which both the representations and the learning mechanism are explicitly provided. 

LLMs might be able to help. We can train them on a suitable corpus and then probe their knowledge. Note that this is not entirely straightforward, for various reasons discussed in \citealt{LanChemlaKatzir:2024a}. Most immediately, while we might be able to directly inspect a grammar within a theory of HLC such as $H_3$ and ask whether it has knowledge of $PROP$, we cannot do the same with current LLMs, whose inner workings are extremely complex and remain mostly opaque, as discussed in Section~\ref{sec:llm:th}. Instead of inspection, we might turn to behavioural measures such as acceptability judgments in minimal pairs (see Section \ref{sec:llm:th} above),  where one member satisfies $PROP$ and is acceptable and a minimal variant violates it and is unacceptable. An immediate concern with this methodology is that, while the theory of HLC under consideration might offer such judgments, current LLMs provide only likelihood assessments, through their distribution over the next token given a prefix. Likelihood assessments are very different from acceptability judgments, as mentioned above. However, if we focus on areas where acceptability and likelihood are reasonably well aligned we might be justified in looking at whether the LLM assigns a higher probability to the $PROP$-satisfying member of the pair than to the $PROP$-violating one (a methodology that has been used in \citealp{MarvinLinzen:2018}, \citealp{Hu:2020} and \citealp{WilcoxLevyFutrell:2019}, among many other works; see Section~\ref{sec:llm:th}). If this probabilistic preference is sufficiently strong, we might consider the fact that the LLM has acquired this preference as reason to think that the learner under our target theory of HLC $H_3$ would acquire the actual $PROP$. This might be so even if the probabilistic preference of the LLM reflects a flawed approximation of $PROP$ rather than knowledge of $PROP$ in any meaningful sense.


Following \citeauthor{WilcoxFutrellLevy:2023} and \citeauthor{LanChemlaKatzir:2024a}, our success criterion considers the model successful if and only if it assigns a higher probability to the grammatical member of the minimal pair. Specifically, we evaluate probabilities on a critical region token whose preceding context is the same in both the grammatical and the ungrammatical sentence (see Table~\ref{tab:phenomena}, where critical regions are indicated with underscores). Note that this is an extremely lenient notion of success: the model is considered successful even if the preference for the grammatical member of the minimal pair is extremely small, and even if ungrammatical sentences are considered more likely than some (less related) grammatical sentences.

\subsection{Methods}

\begin{table}[htbp]
\begin{tabularx}{\textwidth}{@{}X r@{ }l r@{ }l@{}}
    \toprule
    \textbf{Model} & \multicolumn{2}{l}{\textbf{Train dataset size}} & \multicolumn{2}{l}{\textbf{Human equivalent}} \\
    \midrule
    CHILDES LSTM,\ Transformer & 8.6  & million tokens    & 10      & months \\
    BabyLM 10M                & 10   & million tokens    & 1       & year \\
    Wikipedia Transformer     & 90   & million tokens    & 8       & years \\
    BabyLM 100M               & 100  & million tokens    & 9       & years \\
    BERT Base Uncased        & $\approx$ 3.5 & billion tokens    & 320     & years \\
    GPT-2                      & $\approx$ 8   & billion tokens    & 730     & years \\ 
    LLaMA-3.2-3b               & $\approx$ 9   & trillion tokens   & 821,250 & years \\
    \bottomrule
    \end{tabularx}
    \caption{Training data size of the eight language models considered here, and
    the human linguistic experience equivalent to these data sizes.}
    \label{tab:model-training-sizes}
\end{table}

Let us illustrate this kind of reasoning with concrete examples. To do so, we use an empirical setup that consists of eight models of different architectures and training schemes, ranging from the rough equivalent of several months of human linguistic input to hundreds of thousands of years, with multiple models in between, as summarized in Table~\ref{tab:model-training-sizes}.\footnote{We use two models from \citealt{yedetore:2023}, an LSTM and a Transformer, that were trained on the CHILDES corpus of child-directed speech; a Transformer we trained on a subset of English Wikipedia, for which we used one of the large Transformer architectures used in \citealt{yedetore:2023}; the \textit{base-strict} and \textit{base-strict-small} versions of the BabyLM LLMs  from \citet{Babylm:proceedings}, trained on the 100M and 10M BabyLM datasets respectively; the \textit{BERT base uncased} version of BERT \citep{Devlin:2018}, trained on the BooksCorpus and English Wikipedia datasets; OpenAI’s GPT-2 \citep{Radford:2019} and Meta's \textit{LLaMA 3.2-3b} LLaMA model (released September 2024, \citealp{MetaLlama3:2024}). See Appendix~\ref{sec:appendix-aps} for further technical detail. All experimental material can be found at \url{https://osf.io/5zh6q/}.} Human equivalents are based on the assumption by \citet{HartRisley:1995} that children are exposed to around 30,000 words per day, while roughly equating words with model tokens. 

The models were tested on three linguistic phenomena using the very lenient success criterion discussed above (see Table~\ref{tab:phenomena}). The models that are most directly relevant to assessing $H_3$ with respect to any given APS are the smaller ones, which are trained on inputs that roughly match the number of words that a child might be exposed to. If these models prefer the grammatical option over the ungrammatical one in the minimal pairs of the test set for a given phenomenon, this can support the idea that $H_3$ would actually acquire the relevant piece of linguistic knowledge from a developmentally-realistic corpus.\footnote{Recall that we do not ask whether the LLMs actually acquire the linguistic knowledge under consideration, nor do we expect them to.} Conversely, lack of  such a preference can weaken the plausibility that $H_3$ would succeed in acquiring the relevant knowledge from a similar corpus.\footnote{As should be clear, even if there is evidence that $H_3$ could acquire the relevant knowledge from a developmentally-realistic corpus there might still be good reasons to adopt an $H_1$ that is strongly biased in favour of this knowledge. Among other things, $H_1$ might provide a better explanation than $H_3$ for the cross-linguistic pattern. We believe that this is the case for the phenomena discussed in the present section and refer the reader to the literature cited below.}

Two of the models (BERT and GPT-2) were trained on the equivalent of hundreds of years of linguistic input. This is clearly beyond what a human child has been exposed to by the time they show evidence of the relevant linguistic knowledge. We included these models to allow proponents of the Proxy View to use LLMs as broadly adequate proxies for $H_3$ that just happen to be extremely data inefficient. 

We included the much larger LLaMA mainly to see whether current architectures are sensible proxies for $H_3$ for the phenomena at hand. If they cannot show even a slight preference for the correct element in minimal pairs after training on the equivalent of hundreds of thousands of years of linguistic input, then perhaps they are incapable of approximating the pattern under consideration, which would suggest that they are not very good proxies for $H_3$ with respect to this pattern.

\begin{table}[h]
\centering
\small
\renewcommand{\arraystretch}{1.3} 
\setlength{\tabcolsep}{4pt} 
\begin{tabular}{p{3.5cm} p{9.5cm}}  
\toprule
\textbf{Phenomenon} & \textbf{Example (\textcolor{teal}{Grammatical}/\textcolor{coral}{*Ungrammatical})} \\ 
\midrule

\multirow{2}{=}{Across-the-board movement (ATB)}  
& \textcolor{teal}{Which boy did you say that Kim hated and that Mary loved \underline{yesterday}?} \\  
& \textcolor{coral}{* Which boy did you say that Kim hated and that Mary loved \underline{Ann} yesterday?} \\  
\midrule

\multirow{2}{=}{Parasitic gaps (PG)}  
& \textcolor{teal}{I know who John's talking to is going to annoy \underline{soon}.} \\  
& \textcolor{coral}{* I know who John's talking to is going to annoy \underline{you} soon.} \\  
\midrule

\multirow{4}{=}{That-trace effects (TTE)}  
& \textcolor{coral}{* Who did you say that \underline{loves} Sue?} \\
& \textcolor{teal}{Who did you say that \underline{Sue} loves?} \\  
& \textcolor{teal}{Who did you say \underline{loves} Sue?} \\  
& \textcolor{teal}{Who did you say \underline{Sue} loves?} \\  
\bottomrule
\end{tabular}
\caption{Examples of linguistic phenomena with grammatical and ungrammatical sentences. Model probabilities are compared on the underscored critical regions.}
\label{tab:phenomena}
\end{table}

\subsection{Test case: across-the-board movement and parasitic gaps}
\label{sec:atb}

Above we mentioned displacement phenomena, illustrating with the position of the auxiliary in yes-no questions in English. Displacement phenomena have featured in recent discussions of the role of LLMs in evaluating APSs, both for the placement of the auxiliary (see, e.g., \citealp{yedetore:2023}) and for wh-movement (see \citealp{ChowdhuryZamparelli:2018}, \citealp{BhattacharyaSchijndel:2020},  \citealp{Chaves:2020}, \citealp{WarstadtParrishLiuMohananeyPengWangBowman:2020}, \citealp{OzakiYurovskyLevin:2022}, \citealp{WilcoxFutrellLevy:2023}, and \citealp{LanChemlaKatzir:2024a}). The latter will concern us in the current two subsections. Here is a simple illustration:

\begin{exe}
\ex{\label{ex:wh:book}[Which book] did you say that Mary read \usc\ last week?}
\end{exe}

The common analysis of wh-movement in generative theories involves a constituent that is attached in one position and is then re-attached higher up in the structure. In (\ref{ex:wh:book}), the constituent \textit{[which book]} is attached within the embedded clause as a sister to \textit{read} and is then re-attached at or near the root of the matrix clause, which among other things leads to it being pronounced in the beginning of the sentence. 

Some work within generative linguistics (\citealp{PearlSprouse:2013}, \citealp{Phillips:2013b}) suggests that a relatively linguistically-neutral theory cannot account for the full knowledge of wh-movement given the kind of data that children are exposed to, an instance of the APS. This is not a consensus view, however, and some recent work such as \citealt{WilcoxFutrellLevy:2023} suggests that LLMs undermine this APS and that these models show that a linguistically-neutral theory would, in fact, acquire a full knowledge of wh-movement from a developmentally-realistic corpus. Following \citeauthor{LanChemlaKatzir:2024a} let us show how LLMs (used as proxies for an undisclosed linguistically-neutral theory of HLC) might bear on the APS under consideration and why they currently do nothing to undermine it.

Before proceeding, let us reiterate that it is hard to tell whether LLMs are good proxies for the relatively linguistically-neutral $H_3$ that seems to be envisioned in works such as \citealt{WilcoxFutrellLevy:2023} in the absence of at least some specifics about that theory. But if they are, and if we focus on cases where correctness and likelihood are reasonably well-aligned, we could reason that if the LLM overwhelmingly assigns much higher probability to the correct member of each minimal pair, this indicates that there was sufficient information in the training data for the target linguistically-neutral theory $H_3$ to acquire the relevant knowledge. And if the LLM does not perform as well, this suggests that the target theory $H_3$ might also struggle. The former possibility weakens the APS, and the latter strengthens it.

Following \citet{LanChemlaKatzir:2024a} we focus on a family of wh-movement configurations where there is complex interaction between two gaps, in a way that seemingly violates island constraints on syntactic movement.\footnote{See \citealt{Ross:1967}, \citealt{Williams:1977,Williams:1990}, \citealt{Engdahl:1983},    \citealt{Haik:1985},  \citealt{Munn:1992}, \citealt{Postal:1993b}, \citealt{Fox:2000},  \citealt{Nissenbaum:2000},  \citealt{HornsteinNunes:2002}, and  \citealt{Sportiche:2024}, among many others, for many important observations about the relevant phenomena that we will not be able to discuss in the present paper. Our simplified treatment here suffices to illustrate the methodological point about evaluating the Proxy View, and since our conclusions below are negative even for the simplified picture, we see no reason to think that current models will successfully approximate the complexities of these phenomena discussed in the literature. Any attempt to make a positive claim about these phenomena would of course need to address these complexities.} The first such phenomenon is \textit{across-the-board (ATB) movement} in coordinate structures. While extraction from a single coordinate is ungrammatical, as in \ref{atb:badleftextraction} and \ref{atb:badrightextraction}, extraction can become grammatical when it takes place from both coordinates, as in \ref{atb:good}:

\begin{exe}
\ex{\label{atb}
\begin{xlist}
\ex[*]{\label{atb:badleftextraction}[Which book] did you say [that Kim wrote \usc\ last year] and [that Mary read a new novel yesterday]?}
\ex[*]{\label{atb:badrightextraction}[Which book] did you say [that Kim wrote a new novel last year] and [that Mary read \usc\  yesterday]?}
\ex[]{\label{atb:good}[Which book] did you say [that Kim wrote \usc\ last year] and [that Mary read \usc\ yesterday]?}
\end{xlist}
}
\end{exe}

The second nuance of wh-movement we test is when extraction from an island is made possible by the presence of another gap downstream, a phenomenon known as a \textit{parasitic gap (PG)}: 

\begin{exe}
\ex{\label{pg}
\begin{xlist}
\ex[*]{\label{pg:bad}I know who [John's talking to \usc] is going to annoy you soon.}
\ex[]{\label{pg:good}I know who [John's talking to \usc] is going to annoy \usc\ soon.}
\end{xlist}
}
\end{exe}

We test the eight models from Table~\ref{tab:model-training-sizes} on their approximation of across-the-board movement and parasitic gaps using the minimal pair method. We compare sentences that both have an upstream filler -- like \textit{which book} in Example~\ref{atb}, or \textit{who} in Example~\ref{pg} -- and differ in whether they are gapped accordingly.\footnote{The paradigm sentences in \citealt{LanChemlaKatzir:2024a} were generated by separate templates for ATB and PG. Here we unify the templates for both phenomena so that they use the same lexical choices where relevant (e.g., proper names), and by adding more lexical choices. From each template, represented by a context-free grammar, we sampled 10,000 sentence pairs and ran them through each model to get the relevant probability values (see Appendix~\ref{sec:appendix-typ}).} As shown in Figure~\ref{fig:atb-pg-raw}, all models other than LLaMA prefer the ungrammatical over the grammatical members of the minimal pairs in the vast majority of cases. This failure suggests that the linguistically-neutral $H_3$ for which the LLMs are taken to be proxies would not acquire wh-movement from a developmentally-realistic corpus.

\begin{figure}
    \includegraphics[width=\columnwidth]{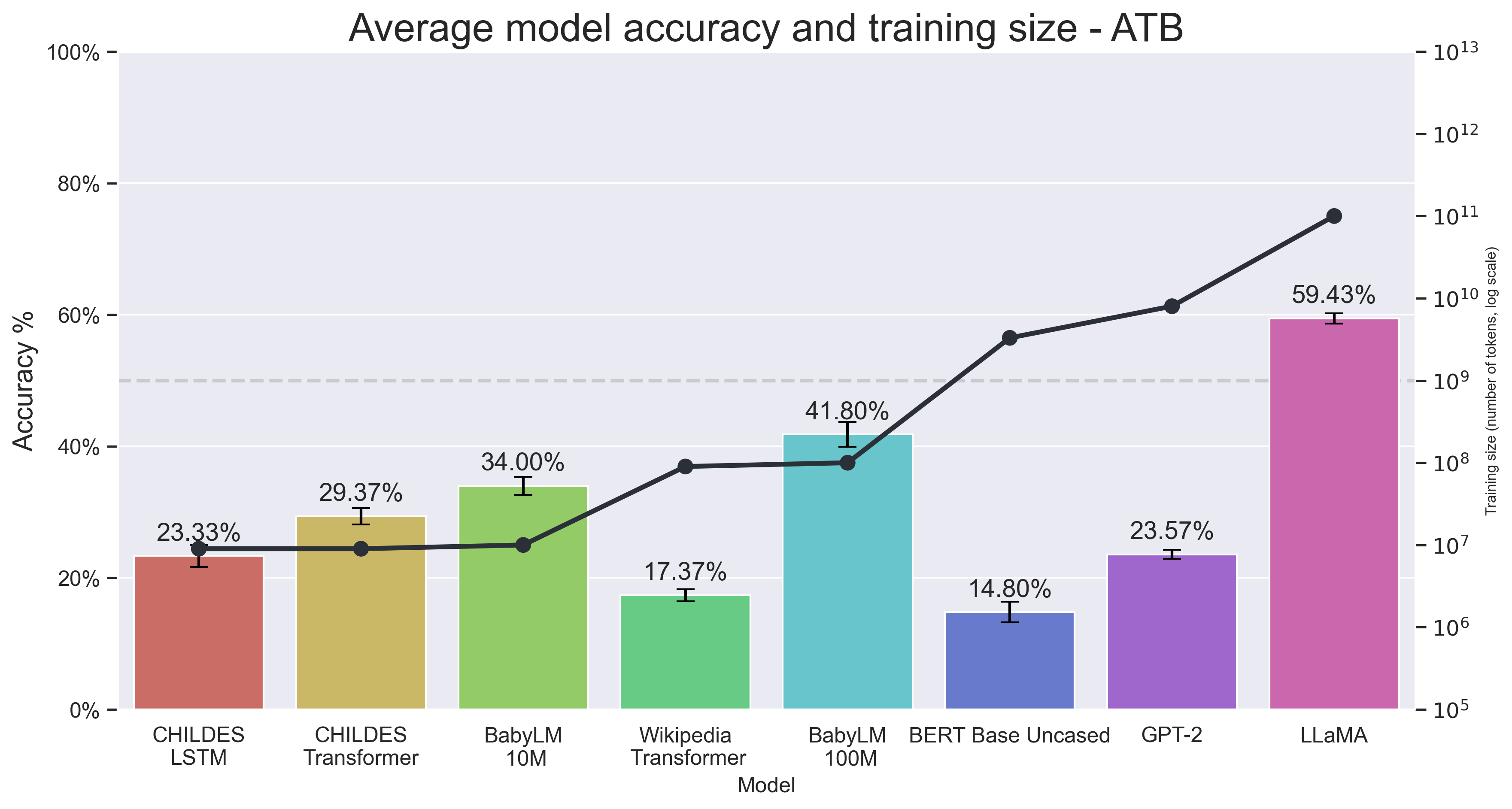} \\[1ex] 
    \includegraphics[width=\columnwidth]{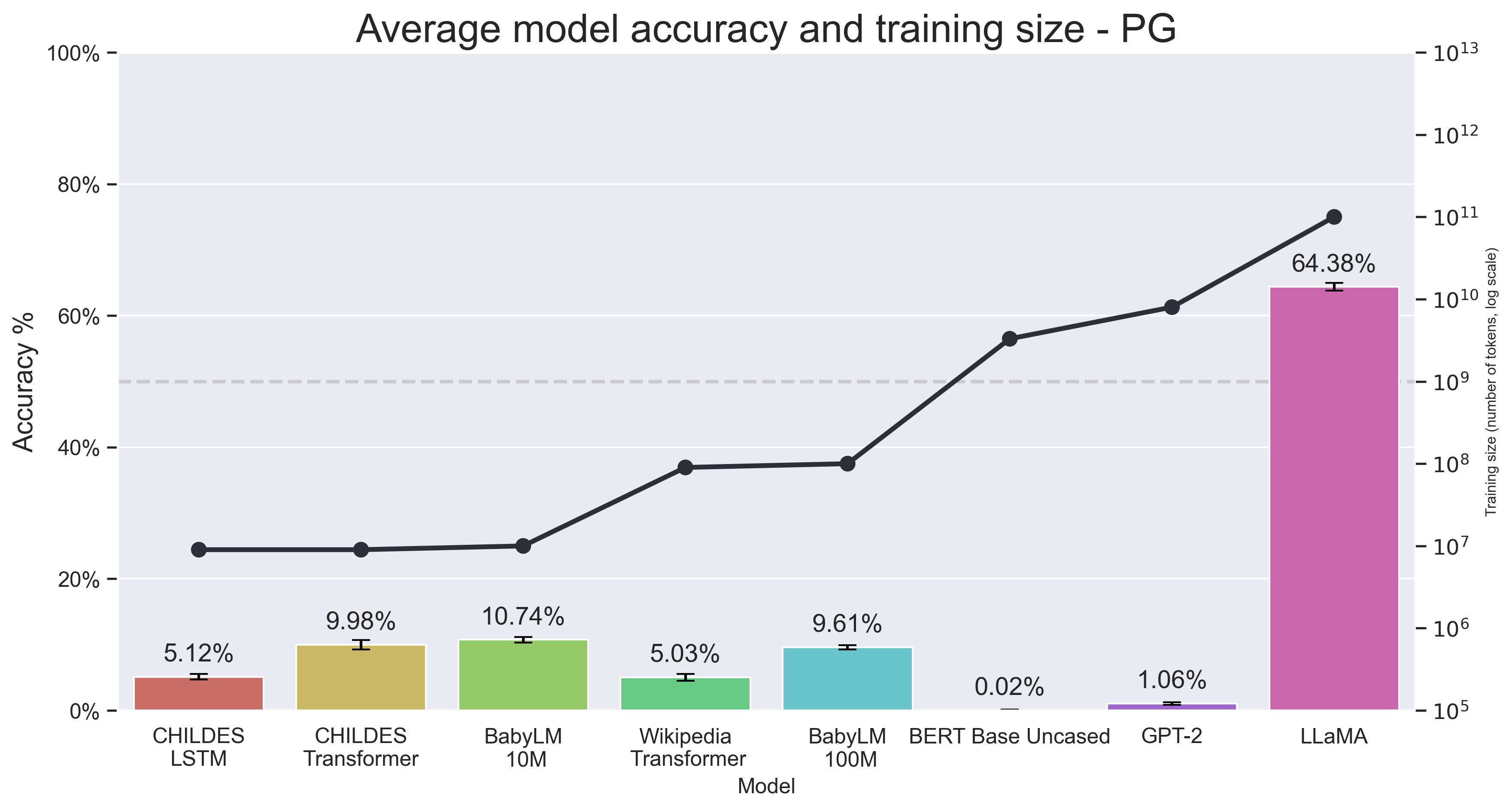}
    \caption{Model accuracy on ATB and PG datasets averaged over five experiment seeds. Accuracy is measured as the proportion of cases where the model assigns a higher probability to the grammatical sentence continuation. The dark plot represents model training sizes (on a logarithmic scale, right).}
    \label{fig:atb-pg-raw}
\end{figure}

\newpage
\subsection{Test case: that-trace effects}
\label{sec:tte}

Another property of wh-movement in English is that when the moved wh-phrase is a subject, the clause from which it moves cannot have an overt complementizer such as \textit{that}:

\begin{exe}
\ex{\label{ex:tte}
\begin{xlist}
\ex[*]{\label{tte:pthatptrace}Who did you say that \usc\ \underline{loves} Sue?_{(+that, +trace)}}
\ex[]{\label{tte:pthatmtrace}Who did you say that \underline{Sue} loves \usc?_{(+that, -trace)}}
\ex[]{\label{tte:mthatptrace}Who did you say \usc\ \underline{loves} Sue?_{(-that, +trace)}}
\ex[]{\label{tte:mthatmtrace}Who did you say \underline{Sue} loves \usc?_{(-that, -trace)}}
\end{xlist}
}
\end{exe}

This phenomenon, known as \textit{that}-trace effects (TTE), has been the topic of much discussion in the linguistic literature.\footnote{See \citealt{Perlmutter:1968}, \citealt{Bresnan:1972}, \citealt{ChomskyLasnik:1977}, and much subsequent work. See \citealt{Pesetksy:2017} for a more recent overview and discussion. As with ATB movement and PG, the present discussion is a simplification, and we refer the reader to the literature for observations and analysis that would be important for any attempt to make positive claims about LLM performance and its potential support for $H_3$.} Building on corpus data from \citealt{PearlSprouse:2013}, \citet{Phillips:2013b} points out that extraction of any kind from clauses with \textit{that} is vanishingly rare in adult-directed corpora and even rarer in child-directed speech: out of 11,308 analyzed utterances from child-directed speech, there were only two instances of non-subject extraction with \textit{that} and zero instances of subject extraction with \textit{that}, contrasting with 159 instances of object extraction without \textit{that}, and 13 instances of subject extraction without \textit{that}. As \citet{Phillips:2013b} notes, these distributional observations strongly suggest an APS for the knowledge behind TTE. Still, one might try to defend an $H_3$ that is not biased in favour of TTE and in which knowledge of TTE is acquired from the input data, perhaps pieced together from several indirect sources of evidence. As with the ATB and PG cases above, we use LLMs as proxies for such a linguistically-neutral $H_3$: if these models adequately approximate TTE, this can be taken to suggest that $H_3$ might succeed in learning TTE, thus weakening the current APS.

We test the same eight LLMs as we did for ATB and PG (see Table~\ref{tab:model-training-sizes}). The basic TTE paradigm in (\ref{ex:tte}) suggests a condition that, through the minimal pair method, reflects an adequate approximation of TTE: given an overt complementizer upstream, the model should prefer an overt subject (\textit{-trace}) to a gap (\textit{+trace}) in the embedded clause. This preference should translate into probabilities so that $P(\text{loves} \mid \text{Who did you say that})$ in \ref{tte:pthatptrace} should be smaller than $P(\text{Sue} \mid \text{Who did you say that})$ in \ref{tte:pthatmtrace}. We thus refer to this condition as $P(\text{+trace} | \text{+that}) < P(\text{-trace} | \text{+that})$. 

\begin{figure}[h]
    \includegraphics[width=\columnwidth]{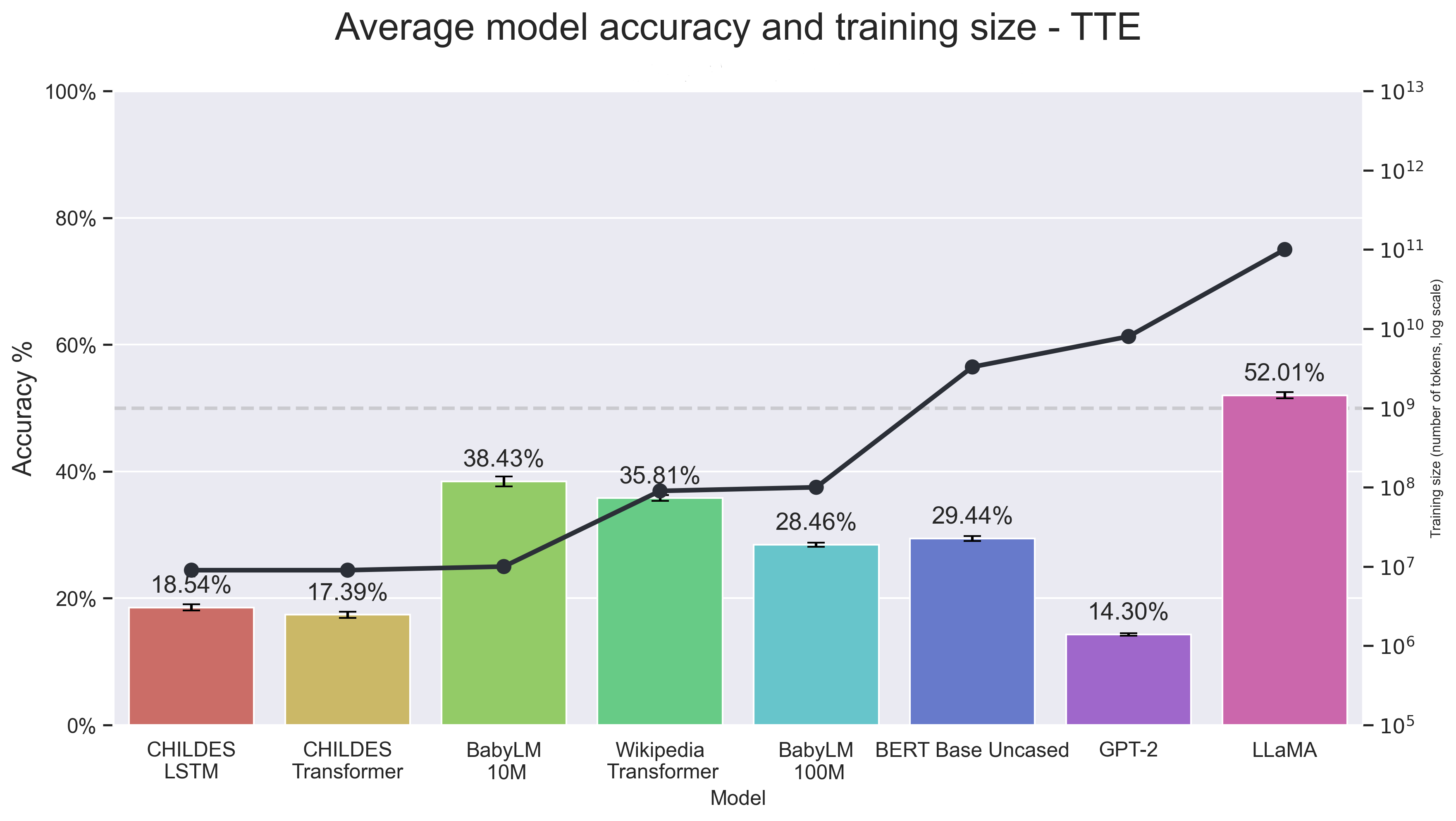} \\[1ex]
    \caption{Model accuracy for the $P(\text{+trace} \mid\text{+that}) < P(\text{-trace} \mid \text{+that})$ criterion over samples of 10,000 sentence pairs from TTE test set. Results are averaged over five seeds. The dark plot represents model training sizes (on a logarithmic scale, right).}
    \label{fig:tte-raw}
\end{figure}

Results for the $P(\text{+trace} | \text{+that}) < P(\text{-trace} | \text{+that})$ criterion are presented in Figure~\ref{fig:tte-raw}. All models except LLaMA overwhelmingly prefer the ungrammatical element to the grammatical one in the minimal pairs of the test. This is very similar to what we saw for ATB and PG above. It is thus hard for us to see how these results can be used to support the ability of $H_3$ to acquire knowledge of TTE from a developmentally-realistic corpus.\footnote{As a reviewer notes, our results are in tension with the apparent ability of more recent versions of ChatGPT to handle TTE (see \citealp{Pesetsky:2024}). In the absence of sufficient detail by OpenAI about its training process we cannot pinpoint the source of this tension, but we note that the training corpora for current versions of GPT are many orders of magnitude bigger than those used for our models, with the possible exception of LLaMA. As the performance of LLaMA reported here suggests, training on sufficiently large corpora (the equivalent of hundreds of thousands of years of linguistic exposure) may suffice for above chance performance by current models. Moreover, it is conceivable that the pretraining of these current versions includes academic papers that explicitly discuss TTE. It is also possible that at least some of the apparent awareness of these versions to TTE comes from the human-guided stages that follow pretraining.}

\section{Large language models and cross-linguistic evidence}
\label{sec:typ}

Consider again the comparison of a strongly linguistically-biased generative theory $H_1$ and a more linguistically-neutral alternative $H_3$. Suppose that the strong linguistic biases of the generative theory $H_1$ can be shown to help derive a cross-linguistic pattern, perhaps by making it easier to represent and learn the languages that adhere to the pattern than those that do not. In the relatively linguistically-neutral theory $H_3$ it can be much harder to see directly whether the attested patterns are predicted to be easier to learn than the unattested ones. One might try to evaluate this matter empirically, by simulating learning of languages of the attested and the unattested kind given the linguistically-neutral theory, but this can be hard to do. Consequently, it can be hard to see whether the theory can explain the cross-linguistic pattern under consideration and whether it can compete with the generative theory. Again, however, if an LLM is a good proxy for $H_3$, the problem is mitigated to some extent. One can examine the ease with which the LLM learns to approximate both languages that follow the pattern under consideration and languages that violate it. If the LLM approximates the former more easily than the latter, this can suggest that the target theory might be able to account for the cross-linguistic pattern (subject to various further assumptions about how learning affects the typology). If, on the other hand, some of the pattern-violating languages are easier to learn than some of the pattern-satisfying ones, the linguistically-neutral theory would need further mechanisms to derive the typology. We outline a preliminary investigation of this kind, building on the setup of \citet{KalliniPapadimitriouFutrellMahowaldPotts:2024}. We discuss limitations of this approach in Section \ref{sec:typ:lim}.

\subsection{Ease of learning by LLM and typological asymmetries}

\citet{KalliniPapadimitriouFutrellMahowaldPotts:2024}\ compare GPT-2 when trained on an English corpus and when trained on corpora that are derived from the original corpus through various perturbations, each perturbed corpus corresponding to an unattested kind of language. \citeauthor{KalliniPapadimitriouFutrellMahowaldPotts:2024} find that the original English corpus was easier for GPT-2 than the derived corpora of languages of unattested kinds on a very particular notion of ease of learning, defined through test set perplexities through time (see Figure~\ref{fig:il-partial-reverse}). In doing so, \citeauthor{KalliniPapadimitriouFutrellMahowaldPotts:2024} can be taken to make use of LLMs as proxies for a linguistically-neutral $H_3$ in the context of explaining various cross-linguistic skews. These skews might initially be taken to support an $H_1$ that is strongly linguistically biased against languages corresponding to the perturbations under consideration. The performance of GPT-2, however, might suggest that a more linguistically-neutral $H_3$ is also compatible with the typology. On this view, languages corresponding to the perturbations might be unattested because of general, non-linguistic biases that are shared by $H_3$ and GPT-2.\footnote{\citet{KalliniPapadimitriouFutrellMahowaldPotts:2024} contextualize their work differently, relating it to the notion of impossible languages \citep{Moro:2016, Musso:2003}. We will not attempt to determine whether the present work is in line with their goals.} 

In the following subsections we extend the setup of \citealt{KalliniPapadimitriouFutrellMahowaldPotts:2024} by adding to their examination of English a few other languages for which the evaluation is simple given current resources. We use corpora for these languages as the basis for the perturbations used by \citeauthor{KalliniPapadimitriouFutrellMahowaldPotts:2024} and for one additional perturbation, listed in Table~\ref{tab:perturbations}. For each perturbation, we use a baseline and a perturbed version for each of the four languages -- English, Italian, Russian, and Hebrew, based on datasets from \citealt{GulordavaBojanowskiGraveLinzenBaroni:2018}. We then train a transformer LLM based on the architecture and training scheme from \citet{yedetore:2023} and consider the validation set perplexities as training progresses.\footnote{The model was trained for 48 hours and validation perplexity was computed every
200 training batches. The 48 hours of training amounted to roughly 1.5 epochs per dataset, with batch size 10, such that each epoch amounted to an average of 181,552 batches. See Appendix~\ref{sec:appendix-typ} for further details.} 

As outlined above, if the setup in \citealt{KalliniPapadimitriouFutrellMahowaldPotts:2024} is intended to support $H_3$, then the behaviour of GPT-2 with respect to their perplexity metric should show that $H_3$ can be part of a successful account of the typology and that the strong linguistic biases of $H_1$ are not necessary. A relatively straightforward argument would be if ease of learning directly matched the typological divide \textit{across languages}, with languages of the attested kind being uniformly easier to learn than languages of the unattested kind. But ease of learning may well be just one out of several interacting factors that give rise to the typology; other plausible factors include communicative efficiency, sociolinguistic pressures, and historical accident. \citeauthor{KalliniPapadimitriouFutrellMahowaldPotts:2024}'s setup does not incorporate such factors explicitly, but one can perhaps consider them as part of the background. Even when such factors are taken into consideration, however, a setup such as \citealt{KalliniPapadimitriouFutrellMahowaldPotts:2024}'s would presumably only be useful if it showed that ease of learning by the relatively linguistically-neutral $H_3$ can play some role in an account of the typology. Perhaps, rather minimally, the \citeauthor{KalliniPapadimitriouFutrellMahowaldPotts:2024} approach could succeed \textit{within languages}, in that the linguistically-unnatural perturbations of each language would be harder to learn than the attested baseline.

As we show, LLM performance does not provide this kind of support to $H_3$. Ease of learning in the sense of \citet{KalliniPapadimitriouFutrellMahowaldPotts:2024} does not match the general typological divide \textit{across languages}, and even \textit{within languages}, the perturbed version is sometimes easier to learn than the attested baseline. From the perspective of the Proxy View, this means that the  relatively linguistically-neutral $H_3$ fails to receive support from the performance of the proxy LLM: if $H_3$ can be part of an explanation of the typological pattern, this is not shown by the current results.

\begin{table}[h] 
    \renewcommand{\arraystretch}{1.5}  
    \setlength{\tabcolsep}{4pt} 
    \begin{tabular}{p{4.5cm}p{10.5cm}}  
        \toprule
        Perturbation & \textcolor{teal}{Attested} / \textcolor{coral}{Perturbed} \\ 
        \bottomrule
        \textit{partial-reverse (Section \ref{partial-reverse-section})} & 
        \textcolor{teal}{colourless\textsubscript{0} green\textsubscript{1} \texttt{<rev>}\textsubscript{2} ideas\textsubscript{3} sleep\textsubscript{4} furiously\textsubscript{5}.} 
        \newline
        \textcolor{coral}{colourless\textsubscript{0} green\textsubscript{1} \texttt{<rev>}\textsubscript{2} furiously\textsubscript{5} sleep\textsubscript{4} ideas\textsubscript{3}.} \\

        \bottomrule
        \textit{full-reverse} (Section \ref{full-reverse-section})& 
        \textcolor{teal}{colourless\textsubscript{0} green\textsubscript{1} \texttt{<rev>}\textsubscript{2} ideas\textsubscript{3} sleep\textsubscript{4} furiously\textsubscript{5}.} 
        \newline
        \textcolor{coral}{Furiously\textsubscript{5} sleep\textsubscript{4} ideas\textsubscript{3} \texttt{<rev>}\textsubscript{2} green\textsubscript{1} colourless\textsubscript{0}.} \\
        \bottomrule
        \textit{switch-indices} (Section \ref{switch-indices-section}) & 
        \textcolor{teal}{colourless\textsubscript{0} green\textsubscript{1} ideas\textsubscript{2} sleep\textsubscript{3} furiously\textsubscript{4}.} 
        \newline 
        \textcolor{coral}{Ideas\textsubscript{2} green\textsubscript{1} colourless\textsubscript{0} sleep\textsubscript{3} furiously\textsubscript{4}.} \\
        \bottomrule
        \textit{token-hop} (Section \ref{token-hop-section}) & 
        \textcolor{teal}{They\textsubscript{0} were\textsubscript{1} sleeping\textsubscript{2} \textbf{v\textsubscript{3}} next\textsubscript{4} to\textsubscript{5} the\textsubscript{6} colourless\textsubscript{7} green\textsubscript{8} ideas\textsubscript{9}.} 
        \newline
        \textcolor{coral}{They\textsubscript{0} were\textsubscript{1} sleeping\textsubscript{2} next\textsubscript{3} to\textsubscript{4} the\textsubscript{5} \textbf{v\textsubscript{6}} colourless\textsubscript{7} green\textsubscript{8} ideas\textsubscript{9}.} \\
        \bottomrule
    \end{tabular}
\caption{Perturbation test cases. Ease of learning is evaluated for the attested (top, green) and perturbed (bottom, orange) versions of English, Italian, Hebrew and Russian.}
\label{tab:perturbations}
\end{table}

\subsection{Test case: \textit{partial-reverse}}
\label{partial-reverse-section}

We demonstrate the use of ease of learning in our first two test cases, in which we consider reversed constructions. In our first test case of this sort -- a cross-linguistic reproduction of the \textit{partial-reverse} perturbation from \citealt{KalliniPapadimitriouFutrellMahowaldPotts:2024} -- 
each sentence is reversed starting from a randomly chosen index in the input sentence, in which a special marker token \textit{\textless rev\textgreater} is inserted:

\begin{exe}
\ex{\label{ex:partial-reverse}
\begin{xlist}
\ex[Baseline:]{\label{partial-reverse-basline} colourless_{0} green_{1} \textit{<rev>}_{2} ideas_{3}  sleep_{4} furiously_{5.}}
\ex[Perturbed:]{\label{partial-reverse-perturbed} colourless_0 green_1 \textit{<rev>}_2 furiously_5 sleep_4 ideas_3.}

\end{xlist}
}
\end{exe}

Reversing linguistic data is generally considered an effective way to generate unnatural data, since it disrupts the language's consistent word orders \citep{MitchellBowers:2020}, and in the case of \textit{partial-reverse}, also unnaturally entangles dependencies involving constituents that cross the \textit{\textless rev\textgreater} token boundary. The target, relatively linguistically-neutral theory $H_3$ in this case might be a version of a generative theory $H_1$ in which there is a parameter for reverse. Any grammar that is statable within $H_3$ now comes in two variants. If the parameter is set to 0, the grammar works as in the original generative theory. But if the parameter is set to 1, every derivation ends with reversing the output starting from a random point. This may or may not be the target theory that \citeauthor{KalliniPapadimitriouFutrellMahowaldPotts:2024} have in mind; other target theories are of course imaginable, and \citeauthor{KalliniPapadimitriouFutrellMahowaldPotts:2024} do not comment explicitly on this. We note that, like \citeauthor{KalliniPapadimitriouFutrellMahowaldPotts:2024}'s other perturbations, reversal seems quite far away from the typological questions that generative linguistics have usually concerned themselves with. Here we stay with \citeauthor{KalliniPapadimitriouFutrellMahowaldPotts:2024}'s perturbations and setup simply in order to illustrate the relevant methodological considerations, but a proper exploration of the Proxy View would need to look at those cross-linguistic patterns that have informed linguistic research.  

\begin{figure}[h]
\includegraphics[width=\columnwidth]{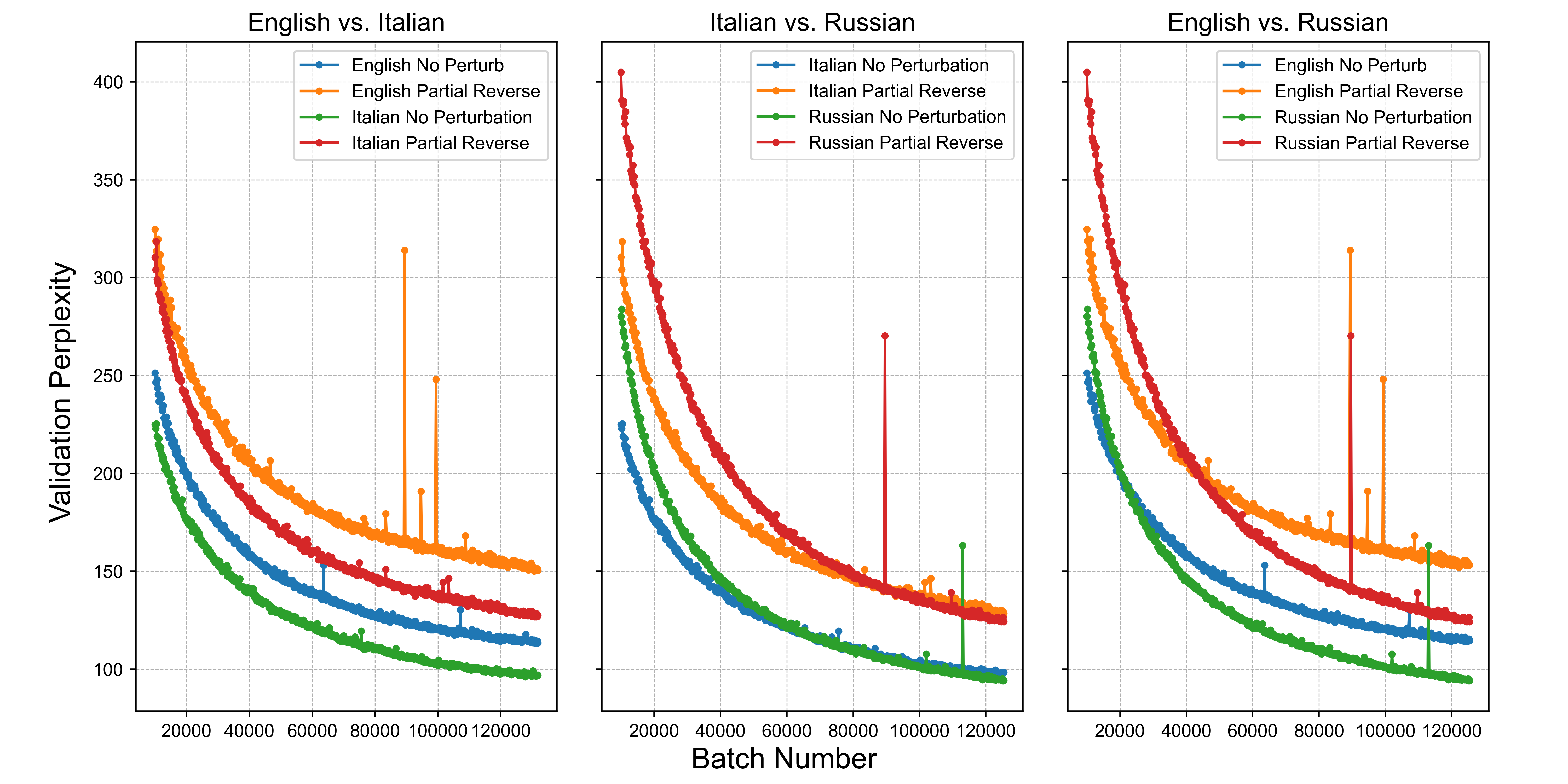}
\caption{Validation perplexity during training for English, Italian, and Russian and their \textit{partial-reverse} perturbations. The results indicate that the learning curves for attested languages are consistently below those of their \textit{partial-reverse} counterparts.}

\label{fig:il-partial-reverse}
\end{figure}

We find that validation perplexities follow an attested-unattested divide \textit{within languages}: the partial-reverse perturbed versions of English, Italian and Russian are harder to learn than their attested counterparts (see Figure~\ref{fig:il-partial-reverse}).\footnote{The \textit{partial-reverse} languages are compared to a modified baseline of English in which the marker \textit{\textless rev\textgreater} is inserted randomly without reversing, to control for the effect for additional textual material in the sentence on perplexities. Since we use the same seed for both perturbations, the markers are inserted in the same indices (see Example \ref{ex:partial-reverse}).}

This, as discussed above, could be part of an argument that linguistic biases are not needed to rule out partial-reverse languages. Consider an $H_3$ that is a serious contender as a theory of HLC, addressing the various arguments regarding HLC in the literature. Suppose that $H_3$ is relatively linguistically neutral in the sense of allowing not just the languages sanctioned by a standard generative $H_1$ but also their partial-reverse versions. An argument in favour of $H_1$ and against $H_3$ might be that these partial-reverse variants are in fact unattested. Our results, following \citealt{KalliniPapadimitriouFutrellMahowaldPotts:2024}, can be taken as a response to this objection: using GPT-2 as a proxy for $H_3$, there is preliminary reason to think that attested languages might be slightly easier to learn than their partial-reverse variants, which might be part of an explanation of why the latter are unattested. This, in turn, could allow us to keep $H_3$ as a contender. 

Other perturbations paint a different picture, as we now show.

\subsection{Test case: \textit{full-reverse}}
\label{full-reverse-section}
 We now consider the perturbation \textit{full-reverse} from \citealt{KalliniPapadimitriouFutrellMahowaldPotts:2024}, where each sentence is reversed in its entirety and a special marker token is randomly inserted: 

\begin{exe}
\ex{\label{ex:full-reverse}
\begin{xlist}
\ex[Baseline:]{\label{full-reverse-baseline} colourless_{0} green_{1} \textit{<rev>}_{2} ideas_{3}  sleep_{4} furiously_{5.}}
\ex[Perturbed:]{\label{full-reverse-perturbed} Furiously_{5} sleep_{4} ideas_{3} \textit{<rev>}_{2} green_{1} colourless_{0}.}

\end{xlist}
}
\end{exe}

\begin{figure}[h]

\includegraphics[width=\columnwidth]{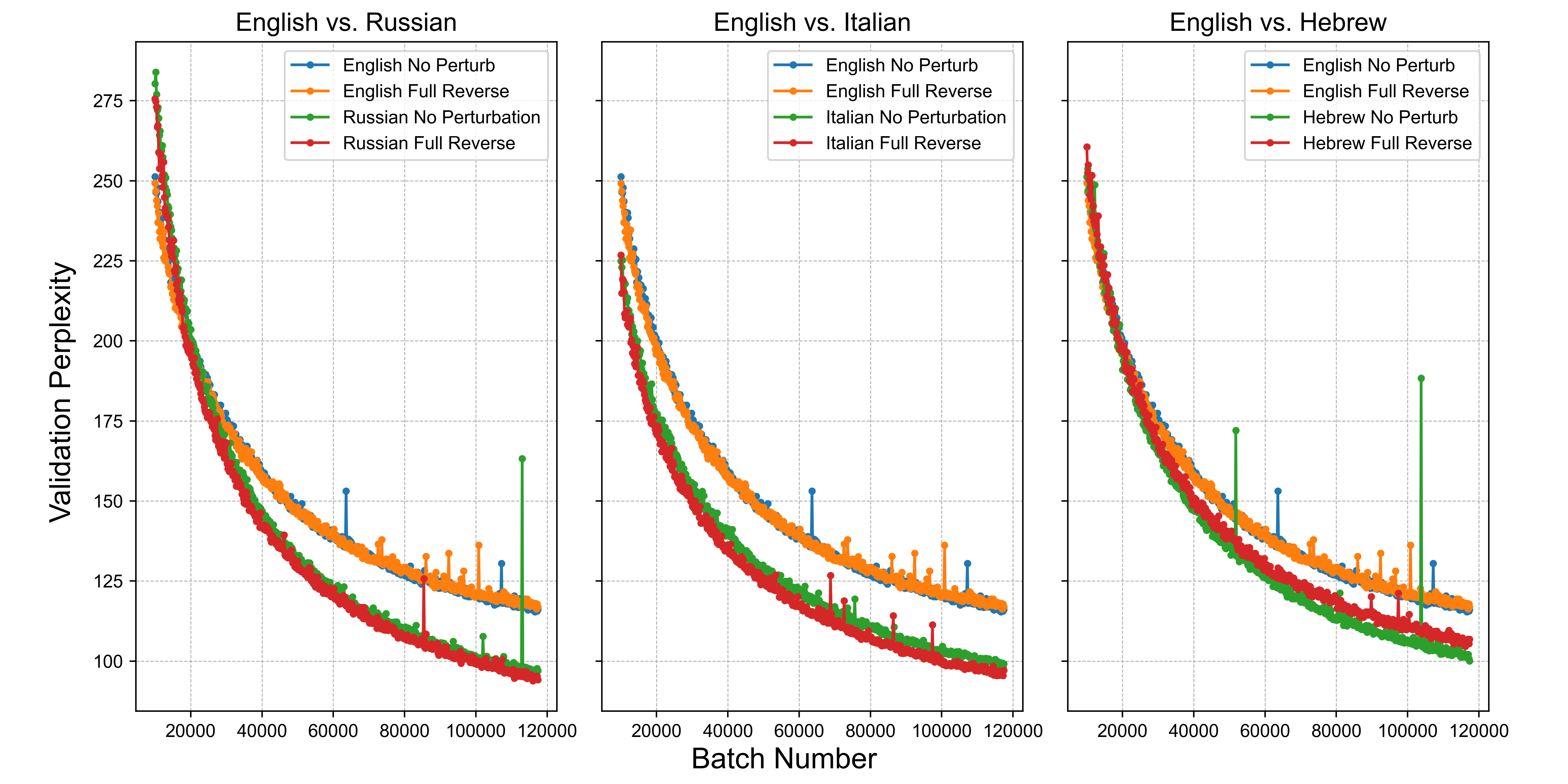}
\caption{Validation perplexity during training for attested (baseline) and \textit{full-reverse} versions of English, Russian, Italian, and Hebrew.}
\label{fig:il-full-reverse}
\end{figure}

We find that \textit{across languages}, the perturbed versions of Italian, Russian and Hebrew are projected as easier than attested English (see Figure~\ref{fig:il-full-reverse}), even though the perturbation is not considered humanly impossible. \textit{Within languages}, Italian is projected as harder than its perturbed version. In this case, ease of learning according to \citealt{KalliniPapadimitriouFutrellMahowaldPotts:2024}’s perplexity metric does not provide support for the idea that the target theory can explain away the typological pattern through asymmetries of ease of learning. 

\subsection{Test case: \textit{switch-indices}}
\label{switch-indices-section}
We turn to another typological asymmetry that the Proxy View does not at present explain. Consider \textit{switch-indices}, where the tokens at index 0 and index 2 in every sentence are switched:

\begin{exe}
\ex{\label{ex:switch-indices}
\begin{xlist}
\ex[Baseline:]{\label{switch-indices-baseline} colourless_{0} green_{1} ideas_{2} sleep_{3} furiously_{4}.}
\ex[Perturbed:]{\label{switch-indices-perturbed} Ideas_{2} green_{1} colourless_{0} sleep_{3} furiously_{4}.}

\end{xlist}
}
\end{exe}

One can again imagine various target theories that allow for languages like English as well as their switch-index variants. And again the question would be whether such a target theory can account for the typological observation that such switch-index variants are unattested.

\begin{figure}[h]
\includegraphics[width=\columnwidth]{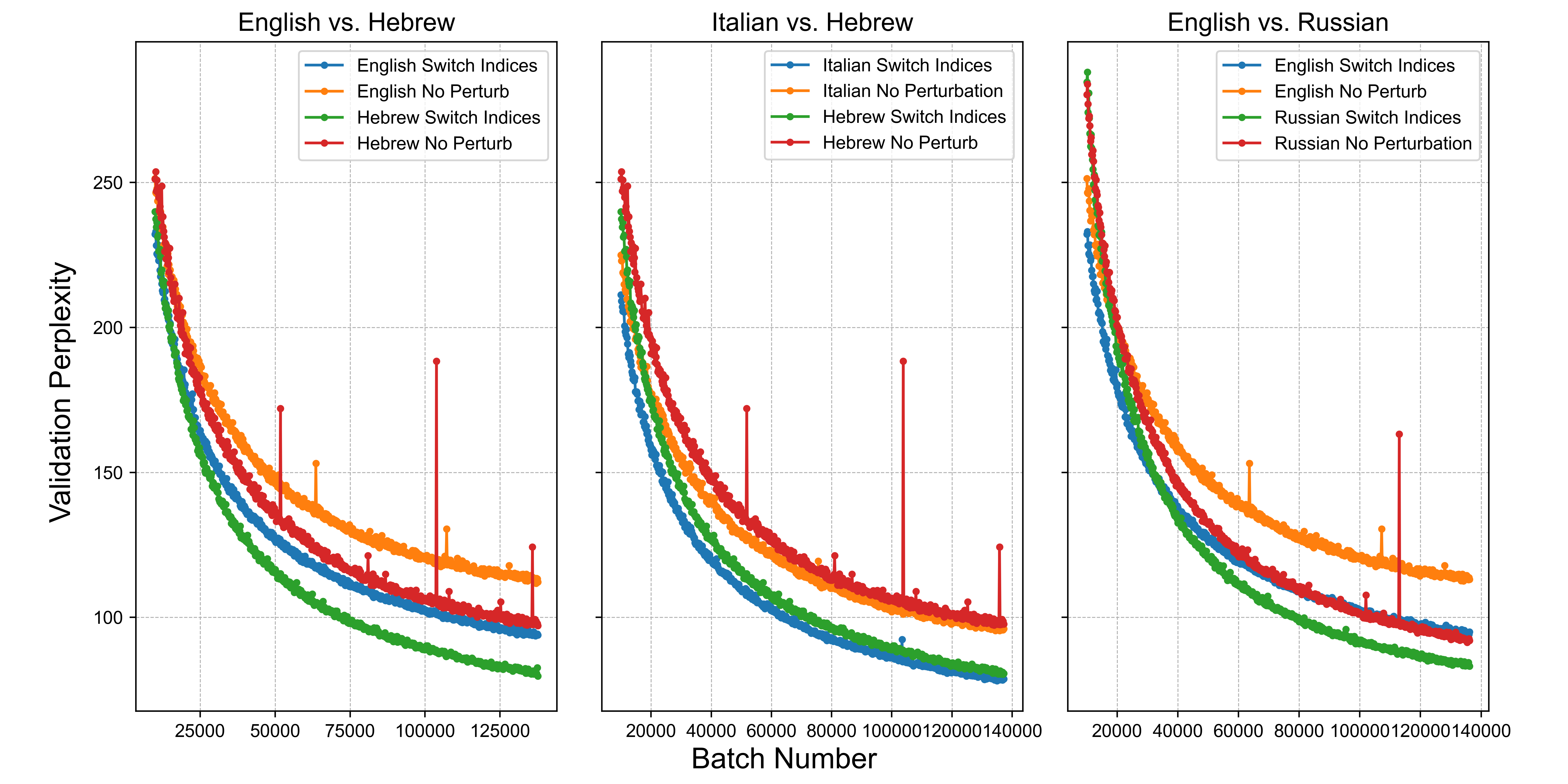}

\caption{Validation perplexity during training for the attested (baseline) and \textit{switch-indices} versions of English, Italian, Hebrew and Russian.}
\label{fig:il-switch-indices}

\end{figure}

If the target theory can account for the typological asymmetry, this is again not detected through the perplexity curve of the LLM proxy: \textit{within languages}, each attested language is wrongly projected as harder than its perturbed version (see Figure \ref{fig:il-switch-indices}).

\pagebreak

\subsection{Test case: \textit{token-hop}, \textit{no-hop}}
\label{token-hop-section}

We look at one final typological asymmetry, based on the \textit{token-hop} perturbation employed by  \citet{KalliniPapadimitriouFutrellMahowaldPotts:2024}.  For each of English, Italian, and Russian, we create a perturbed variant by inserting a new marker token three tokens after each verb:\footnote{New marker tokens were randomly generated as single-character tokens that do not already exist in the model's vocabulary (English: ``v'', Italian: ``v'', Russian: ``\textit{\foreignlanguage{russian}{Ч}}'').}

\begin{exe}
\ex{\label{ex:token-hop-no-hop}
\begin{xlist}
\ex[no-hop (baseline):]{\label{no-hop-baseline} They_0 were_1 sleeping_2 \textbf{v_3} next_4 to_5 the_6 colourless_7 green_8 ideas_9.}
\ex[token-hop (perturbed):]{\label{token-hop-perturbed} They_0 were_1 sleeping_2 next_3 to_4 the_5 \textbf{v_6} colourless_7 green_8 ideas_9.}

\end{xlist}
}
\end{exe}

\begin{figure}[h!]
\includegraphics[width=\columnwidth]{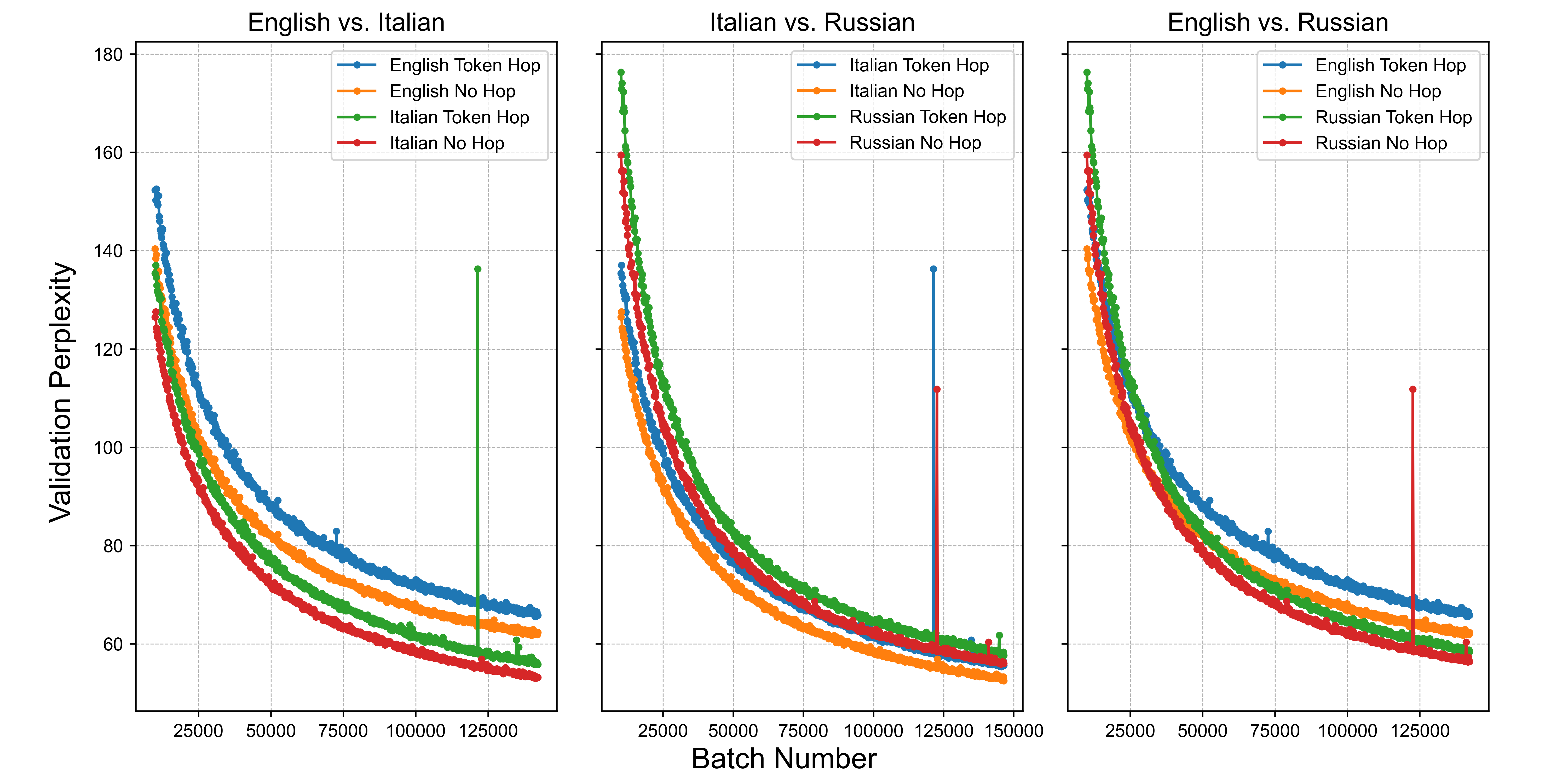}
\caption{Validation perplexity during training for the \textit{no-hop} (baseline) and \textit{token-hop} versions of English, Italian, and Russian.}
\label{fig:il-token-hop}

\end{figure}

Since additional textual material is inserted, in a way that might confound perplexities over time, we compare the perplexities achieved by our model on \textit{token-hop} datasets to \textit{no-hop} datasets, in which the marker token is inserted exactly after each verb. This  controls for the effect of the marker token itself, such that difference in perplexities can be attributed solely to the effect of the token-counting generalization. The \textit{no-hop} version adheres to attested generalizations, as it can be thought of as a post-verbal clitic that marks for parts of speech. Learning \textit{token-hop} on the other hand crucially requires counting tokens, an ability that is particularly non-humanlike, because rules that require counting linear positions in a string violate the strictly hierarchical nature of linguistic generalizations across languages. Although \textit{within languages} the distance of the token from its associated verb has an effect on perplexity, \textit{across languages} we find that the \textit{no-hop} version of English is still harder to learn than \textit{token-hop} Italian and Russian (see Figure~\ref{fig:il-token-hop}).  Unperturbed English datasets are consistently projected as harder than \textit{token-hop} perturbed versions of Italian and Russian, despite a clear typological skew against counting rules. The same point is made with the validation perplexities of \textit{token-hop} Italian and \textit{no-hop} Russian patterning very similarly.


\subsection{Limitations of the typology experiment}
\label{sec:typ:lim}

In this section, we have considered ease of learning by an LLM as a proxy for a linguistically-neutral theory that can explain typological patterns. 
It should be emphasized that the inferences from this argument are particularly weak. On the one hand, even if the linguistically-neutral theory $H_3$ derives a learning asymmetry between pattern-satisfying and pattern-defying languages, this asymmetry might not suffice to account for a robust typological pattern. On the other hand, even if the linguistically-neutral theory $H_3$ does not derive a learning asymmetry (as was the case with respect to several perturbations above), it might still be able to account for it in terms of other factors, such as communicative pressure. In order to overcome this, one could try to situate the learning component within a model of cultural evolution, which allows for the amplification of small asymmetries over generations (see \citealp{KirbyDowmanGriffiths:2007},  \citealp{NiyogiBerwick:2009}, and \citealp{BrochhagenFrankeRooij:2018}, among others) and can potentially balance multiple interacting pressures. We believe that such a model is the proper context for the use of ease of learning by LLMs to evaluate typological asymmetries, but following \citeauthor{KalliniPapadimitriouFutrellMahowaldPotts:2024}, whose setup we use here, we leave such a model for future work and only assess currently available models. All we can conclude at present is that, with the possible exception of partial reverse, \citeauthor{KalliniPapadimitriouFutrellMahowaldPotts:2024}'s setup does not currently provide support to the undisclosed $H_3$ on the Proxy View.\footnote{For further discussion of \citealt{KalliniPapadimitriouFutrellMahowaldPotts:2024} and its limitations see \citealt{Hunter:2025}.}

\section{Conclusion}
\label{sec:conc}

LLMs have been presented as a challenge to more traditional approaches to the study of HLC, and in particular to generative linguistics. A blunt version of this idea is the LLM Theory, promoted by \citet{Piantadosi:2023}, which maintains that current LLMs are themselves good theories of HLC. Current LLMs can, of course, be viewed as theories of HLC, but this treatment does these models no favours, as the literature was quick to note. 

A more cautious approach is the Proxy View, possibly the position of \citet{WilcoxFutrellLevy:2023} and  \citet{MahowaldIvanovaBlankKanwisherTenenbaumFedorenko:2024}, which maintains that current LLMs are good proxies -- as far as various behavioural measures are concerned -- for good theories of HLC that are more linguistically-neutral than generative theories. No commitment to shared essentials between LLMs and the relevant good theories is implied. The Proxy View is of course perfectly coherent, but in contrast to the LLM Theory, it is currently too vague about the theories of HLC that it champions to be of much direct use within empirical science. Empirical science is based on inference to the best explanation and proceeds through competition between reasonably well-understood theories; saying just that one's favourite theory learns roughly as well as an LLM is insufficient for this purpose. One would like to see enough detail at least for a proper comparison of the relevant theory to generative linguistics' theories in light of the discoveries and considerations in the literature of the past 70 years or so. More than anything else, the present paper is a plea for supporters of the Proxy View to provide this kind of detail and evaluation for the theory they have in mind.

Still, already at this stage we can outline an empirical evaluation of the Proxy View on various kinds of test cases, and here we focused on two: alignment with the stimulus and cross-linguistic variation. The results were not encouraging for the Proxy View. The LLMs failed to even approximate key pieces of knowledge that humans have, even when the models were trained on much larger corpora than children are exposed to. The LLMs also had an easier time approximating various languages of typologically-unattested kinds than actual human languages, at least with respect to a success criterion used in \citet{KalliniPapadimitriouFutrellMahowaldPotts:2024}. We conclude that, at least as far as the models and phenomena discussed here are concerned, current LLMs do nothing to suggest changes to linguistic theory.

\appendix 
\section{Appendix: experiment details}
\label{sec:appendix-aps}

\subsection{General Notes}
In all APS experiments, probabilities are evaluated on the critical region in boldface (as in the example sentences in Tables~\ref{tab:cfg-atb}, \ref{tab:cfg-pg}, \ref{tab:cfg-tte}). For models listed as unidirectional, the probability value takes into consideration the left context of the evaluated token, whereas for bidirectional models, the probability value factors in both the left and the right context. The bidirectional models we consider were pretrained with masked language modelling (\textit{fill-mask}). 

In all generated sentences, the critical region unambiguously disambiguates the sentence as an instance of the phenomenon under consideration (there is no other grammatical reading). To promote lexical variation in target regions, possible target tokens were chosen at random from a POS-tagged English Wikipedia dump. We filtered out potential target tokens that were not part of the vocabulary of one the models considered in our setup, which are detailed in Table~\ref{tab:models-verbose}.

\begin{table}[h!]
\centering
\resizebox{\columnwidth}{!}{%

\begin{tabular}{l|lll}
\hline
\textbf{Model Name} & \textbf{Training Size} & \textbf{Directionality} & \textbf{Source} \\
\hline
CHILDES LSTM & 8.6M & Uni & \citealt{yedetore:2023} \\
CHILDES Transformer & 8.6M & Uni & \citealt{yedetore:2023} \\
BabyLM 10M (\textit{strict-small}) & 10M & Bi & \citealt{Babylm:proceedings}  \\
Wikipedia Transformer & 90M & Uni & Trained by authors \\
BabyLM 100M & 100M & Bi & \citealt{Babylm:proceedings} \\
BERT Base Uncased & $\approx$3.5B & Bi & \citealt{Devlin:2018} \\
GPT-2 & $\approx$8B & Uni & \citealt{Radford:2019} \\
LLaMA 3.2-3b & $\approx$9T & Uni & \citealt{MetaLlama3:2024} \\
\hline
\end{tabular}
}
\caption{Details of the models introduced in Table~\ref{tab:model-training-sizes} and used in the experiments in Section~\ref{sec:aps}.}
\label{tab:models-verbose}
\end{table}

\subsection{ATB and PG}
For the ATB and PG experiments, we replicate the experimental setup from \citealt{LanChemlaKatzir:2024a} with a wider variety of test sentences to increase robustness. We also unify the templates for both phenomena
so that they use the same lexical choices where relevant (e.g., proper names). From each template, represented by a context free grammar, we generated tuples of sentences that correspond to the \textit{(+filler,+gap), (+filler,-gap)} conditions (see CFGs and example sentences in Tables~\ref{tab:cfg-atb} and~\ref{tab:cfg-pg}). We then evaluated the success criterion $P(+filler,-gap) < P(+filler,+gap)$ on a sample of 10,000 tuples for each phenomenon. 
Underlined words alternate according to the $\pm filler$ condition; words marked in boldface in Tables~\ref{tab:cfg-atb} and~\ref{tab:cfg-pg} denote the position where the $\pm gap$ condition becomes evident and probability is measured. We evaluate the performance of all eight models from Table~\ref{tab:models-verbose}, including bidirectional ones. To render the \citeauthor{LanChemlaKatzir:2024b} setup viable for testing bidirectional sentences we added a spillover region at the end of every sentence, following a reviewer's suggestion. 
\begin{table}[h!]
    \begin{tabular}{p{9cm}}
        \hline
        ATB Grammar \\
        \hline
        \\
$S \rightarrow \angles{PREAMBLE}  \angles{F} \angles{\pm G}\ \angles{SPILLOVER} $
\\
$\langle PREAMBLE \rangle \rightarrow \ \textit{I know}\ \angles{WH} \angles{EMBEDDER}$
\\
$\angles{WH} \rightarrow \text{`which boys'}\ |\ \text{`which girls'}$
\\
$\angles{EMBEDDER} \rightarrow \text{`you think'}\ |\ \text{`we believe'}$
\\
$\angles{+F} \rightarrow \ \text{that}\ \angles{NAME1} \angles{VP1} \angles{ADV1}$
\\
$\angles{+G} \rightarrow \angles{LINK}\ \angles{NAME2} \angles{VP2}\angles{\textbf{ADV2}}$
\\
$\angles{-G} \rightarrow \angles{LINK}\ \angles{NAME2} \angles{VP2}\ \angles{\textbf{OBJ}}\ \angles{ADV2}$
\\
$\angles{NAME1} \rightarrow \text{`Bob'}\ |\ \text{`John'}$
\\
$\angles{NAME2} \rightarrow \text{`Mary'}\ |\ \text{`Jennifer'}$
\\
$\angles{LINK} \rightarrow \text{`and that'}$
 \\ 
$\angles{ADV1} \rightarrow \text{`shortly'}\ |\ \text{`eventually'}$
 \\ 
$\angles{ADV2} \rightarrow \text{`soon'}\ |\ \text{`today'}\ |\ \text{`now'}$
 \\ 
$\angles{VP1} \rightarrow \text{`will meet'}\ |\ \text{`will see'}$
 \\ 
$\angles{VP2} \rightarrow \ \text{`will hug'}\ |\ \text{`will slap'}\ |\ \text{`will kiss'}$
 \\ 
 $\angles{OBJ} \rightarrow \text{`you'}\ |\ \text{`us'}\ |\ \text{`Kim'}$
\\
$\angles{SPILLOVER} \rightarrow \text{`or some other time' | `or next week'}$
\\
$\cdots$
\\

$\Rightarrow$ I know \underline{which boys} you think that John will meet shortly and that Mary will hug \textbf{soon} \textit{or some other time}. $_{(+filler, +gap)}$
\\
$\Rightarrow$ *I know \underline{which boys} you think that John will meet shortly and that Mary will hug \textbf{you} soon \textit{or some other time}. $_{(+filler, -gap)}$
\\
\\
\hline
\hline \\
    \end{tabular}
    \caption{
    Excerpt from the context-free grammar used to generate Across-the-Board paradigms for the experiments in Section~\ref{sec:aps}. Critical regions are indicated in boldface. Spillover regions relevant only for bidirectional models are indicated in italics.}
    \label{tab:cfg-atb}
\end{table}

\begin{table}[h!]
\begin{tabular}{p{9.5cm}}
\hline
PG Grammar \\
\hline
\\
\\
$S \rightarrow \angles{PREAMBLE} \angles{\pm F} \angles{\pm G} \angles{SPILLOVER}$
\\
$\langle PREAMBLE \rangle \rightarrow \ \textit{I know}$
\\
$\angles{+F} \rightarrow \ \textit{\underline{who}}\  \langle NAME1 \rangle \angles{GEN} \angles{NP} $
\\
$\angles{-F} \rightarrow \ \textit{\underline{that}}\  \langle NAME1 \rangle \angles{GEN} \angles{NP} \angles{\underline{NAME2}} $
\\
$\angles{+G} \rightarrow \angles{LINK}\angles{V} \angles{\textbf{ADV}}$
\\
$\angles{-G} \rightarrow \angles{LINK}\angles{V} \angles{\textbf{OBJ}}\angles{ADV} $
\\
$\angles{GEN} \rightarrow \ \textit{'s} $
\\
$\angles{NP} \rightarrow \angles{NP\_SIMPLE}\ |\ \angles{NP\_COMPLEX}$
 \\ 
$\angles{NP\_SIMPLE} \rightarrow \angles{GERUND}$
 \\ 
$\angles{NP\_COMPLEX} \rightarrow \angles{N\_EMBEDDED}\ \text{`to'}\ \angles{V\_EMBEDDED}$
 \\ 
$\angles{LINK} \rightarrow \text{`is about to'}\ |\ \text{`is likely to'}\ |\ \text{`is going to'}\ |\ \text{`is expected to'}$
 \\ 
$\angles{V} \rightarrow \text{`bother'}\ |\ \text{`annoy'}\ |\ \text{`disturb'}$
 \\ 
 $\angles{OBJ} \rightarrow \text{`you'}\ |\ \text{`us'}\ |\ \text{`Kim'}$
\\
$\angles{GERUND} \rightarrow \text{`talking to'}\ |\ \text{`dancing with'}\ |\ \text{`playing with'}$
 \\ 
$\angles{N\_EMBEDDED} \rightarrow \text{`decision'}\ |\ \text{`intent'}\ |\ \text{`effort'}\ |\ \text{`attempt'}\ |\ \text{`failure'}$
 \\ 
$\angles{V\_EMBEDDED} \rightarrow \text{`talk to'}\ |\ \text{`call'}\ |\ \text{`meet'}\ |\ \text{`dance with'}\ |\ \text{`play with'}$
 \\ 
$\angles{ADV} \rightarrow \text{`soon'}\ |\ \text{`eventually'}$
\\
$\angles{SPILLOVER} \rightarrow \text{'or some other time' | 'or next week'}$ \\
$\cdots$
\\

$\Rightarrow$ I know \underline{who} John's talking to is going to annoy \textbf{soon} \textit{or some other time}. $_{(+filler, +gap)}$
\\
$\Rightarrow$ * I know \underline{who} John's talking to is going to annoy \textbf{you} soon \textit{or some other time}. $_{(+filler, -gap)}$
\\
\\
\hline
\hline \\
\end{tabular}
\caption{Excerpt from the context-free grammar used to generate Parasitic Gap paradigms for the experiments in Section~\ref{sec:aps}. Critical regions are indicated in boldface. Spillover regions relevant only for bidirectional models are indicated in italics.}
\label{tab:cfg-pg}
\end{table}

\subsection{TTE}

To create test sets, we generate 1,178,496 sentence pairs, out of which
we randomly sample 10,000 pairs per experiment. Paradigms are varied lexically in several ways: we first create paradigms that differ
structurally from each other, to control for the possibility that our results are an
artifact of the specific sentence structure introduced in the example sentences in Table~\ref{tab:cfg-tte}. We
also introduce lexical variation to all non-target elements, such as the main subject,
the embedding verb, the wh-element and the verb inside the wh-clause. Most
importantly, lexical choices for target verbs and nouns are varied using 30 target nouns and 30 target verbs. We perform five inference experiments on five such samples and consider average accuracy over all experiments. Pairs are evaluated with the success criterion introduced in Section~\ref{sec:tte}.

\begin{table}[h!]
\resizebox{\columnwidth}{!}{%
\begin{tabular}{p{13cm}}\hline
TTE Grammar \\
\hline
\\
\\
$S \rightarrow \angles{PREAMBLE} \angles{F} \angles{EMBEDDING\_VERB} \angles{COMP} \angles{TARGET\_SUBJ} \angles{TARGET\_VERB}$
\\
$\angles{PREAMBLE} \rightarrow \text{'He knows'}\ |\ \text{'The boy asked'}\ |\ \text{'The girl knew'} \ldots
$
\\
$\angles{F} \rightarrow \text{who}\ $
\\
$\angles{EMBEDDING\_VERB} \rightarrow \text{`think'}\ |\ \text{`believe'}\ |\ \text{`say'}$
\\
$\angles{COMP} \rightarrow \text{that}\ |\ \varepsilon$
\\
$\angles{TARGET\_SUBJ} \rightarrow \text{`people'}\ |\ \text{`children'}\ |\ \text{`parents'}\ |\ \text{`scientists'}\ |\ \text{`kids'} \ldots$  
\\
$\angles{TARGET\_VERB} \rightarrow \text{'are'}\ |\ \text{'have'}\ |\ \text{'ate'}\ |\ \text{'got'}\ |\ \text{'saw'}\ |\ \text{'met'} \ldots$
\\
$\cdots$
\\

$\Rightarrow$ * He knows who you think that \textbf{have}
children $_{(+that, +trace)}$
\\
$\Rightarrow$ He knows who you think that \textbf{children} have $_{(+that, -trace)}$
\\
$\Rightarrow$ He knows who you think \textbf{have} children $_{(-that, +trace)}$
\\
$\Rightarrow$ He knows who you think \textbf{children} have $_{(-that, -trace)}$
\\
\\
\hline
\hline
\end{tabular}
}
\caption{Excerpt from the context-free grammar used to generate That-Trace-Effect paradigms for the experiments in Section~\ref{sec:aps}.}
\label{tab:cfg-tte}
\end{table}

\section{Appendix: datasets and model building}
\label{sec:appendix-typ}

\subsection{Dataset Creation}
\label{dataset-creation}
Our baseline datasets are based on Wikipedia dumps in English, Italian, Hebrew and Russian from \citet{GulordavaBojanowskiGraveLinzenBaroni:2018}, who extracted 90M token subsets for each language and switched all tokens that do not belong to the top 50K most frequent tokens into $\angles{unk}$ tokens (see \citealt{GulordavaBojanowskiGraveLinzenBaroni:2018}). For each baseline dataset we create perturbed \textit{train}, \textit{test} and \textit{validation} datasets (see Table~\ref{tab:perturbations}), similarly to \citealt{KalliniPapadimitriouFutrellMahowaldPotts:2024}, such that each perturbed dataset has the same number of tokens as its corresponding baseline dataset. 

The \textit{no-hop} and \textit{token-hop} perturbations require POS tagging, which we performed using the \textit{spaCy} Python library. Since the POS tagging accuracy is low for Hebrew, we did not perform \textit{no-hop} and \textit{token-hop} experiments on it. New marker tokens were randomly generated as single-character tokens that do not already exist in the model's vocabulary (English: ``v'', Italian: ``v'', Russian: ``\textit{\foreignlanguage{russian}{Ч}}'').\\We used the following spaCy NLP pipelines to perform POS tagging and identify verbs: 
\begin{enumerate}
    \item ``en\_core\_web\_sm'', accuracy of POS tagger: 0.97
    \item ``it\_core\_news\_sm'', accuracy of POS tagger: 0.97
    \item ``ru\_core\_news\_sm'', accuracy of POS tagger: 0.99
\end{enumerate}
We chose to replace the 3rd person agreement \textit{token-hop} used in \citet{KalliniPapadimitriouFutrellMahowaldPotts:2024} with a POS marker because of the lower accuracy of morphological taggers on languages that are not English and cross-linguistic differences in verb morphology.

\subsection{Model Training and Evaluation}

We use the optimal Transformer architecture chosen in a hyperparameter search conducted by \citealt{yedetore:2023}, which has four layers, a hidden and embedding size of 800, a batch size of 10, a dropout rate of 0.2 and a learning rate of 5. Every 200 batches, we evaluate the average perplexity per token of the current trained model on a held out validation set. We set a 48-hour training threshold, during which each model processed a different number of batches. For consistency in comparison graphs, we use the smallest batch count achieved and compare all models up to that point. Table~\ref{tab:typ-batches} details the number of training batches achieved by the Transformer model on each perturbed dataset.

\begin{table}[h!]
\
\renewcommand{\arraystretch}{1.3}  

\begin{tabular}{p{1.7cm}|p{3cm}p{3cm}p{1.5cm}}
\hline

\textbf{Language} & \textbf{Perturbation} & \textbf{Size of validation set (batches)} & \textbf{Training batches} \\
\hline
\multirow{6}{*}{English} & \textit{no-perturb} & 138430 & 137600 \\
                         & \textit{full-reverse}   & 143518 & 117400 \\
                         & \textit{partial-reverse}   &  143517 & 131800 \\
                         & \textit{switch-indices}   & 166116 & 205516 \\
                        & \textit{no-hop}   & 181502 &  189102 \\
                         & \textit{token-hop} &  181502   & 142200 \\
\hline
\multirow{6}{*}{Italian} & \textit{no-perturb} & 138589 & 137200 \\
                         & \textit{full-reverse}   & 143840 & 154200 \\
                         & \textit{partial-reverse}   & 143506 & 152000 \\
                         & \textit{switch-indices}   & 166307 & 213507 \\
                         & \textit{no-hop}   & 181552 & 146400 \\
                         & \textit{token-hop}  & 181552 & 191752 \\
\hline
\multirow{6}{*}{Russian} & \textit{no-perturb} & 138602 & 136200 \\
                         & \textit{full-reverse}   & 145993 & 153600 \\
                         & \textit{partial-reverse}   & 145984 & 125400 \\
                         & \textit{switch-indices}   & 166323 & 213323 \\
                         & \textit{no-hop}   & 178771 & 148800 \\
                         & \textit{token-hop}   & 178771 & 192571 \\

\hline
\multirow{6}{*}{Hebrew} & \textit{no-perturb}& 137773 & 138000 \\
                         & \textit{full-reverse}   & 142884 & 154800 \\
                         & \textit{partial-reverse}   & 171460 & 203860 \\
                         & \textit{switch-indices}   & 165328 & 210128 \\
                         & \textit{no-hop (see \ref{dataset-creation})}   & - & - \\
                         & \textit{token-hop (see \ref{dataset-creation})}   & - & - \\
\hline
\end{tabular}
\caption{Number of training batches achieved in 48 hours of training on our perturbed datasets from Section~\ref{sec:typ}.}
\label{tab:typ-batches}
\end{table}

\clearpage
\bibliographystyle{plainnat}
\bibliography{sources}

\end{document}